\documentclass{article}

\usepackage{arxiv}

\usepackage[utf8]{inputenc} 
\usepackage[T1]{fontenc}    
\usepackage{hyperref}       
\usepackage{url}            
\usepackage{booktabs}       
\usepackage{amsfonts}       
\usepackage{nicefrac}       
\usepackage{microtype}      
\usepackage{lipsum}
\usepackage{epsfig}
\usepackage{graphicx}
\usepackage{amsmath}
\usepackage{amssymb}
\usepackage{math}
\usepackage{booktabs, multirow}  

\title{Person Re-identification with  Adversarial Triplet Embedding}

\author{
Xinglu Wang\\
College of Information Science \& Electronic Engineering\\
Zhejiang University, China\\
\texttt{xingluwang@zju.edu.cn}\\
}

\begin{document}
\maketitle

\begin{abstract}
  Person re-identification is an important task and has widespread applications in video surveillance for public security. In the past few years, deep learning network with triplet loss has become popular for this problem. However, the triplet loss usually suffers from poor local optimal and relies heavily on the strategy of hard example mining. In this paper, we propose to address this problem with a new deep metric learning method called Adversarial Triplet Embedding (ATE), in which we simultaneously generate adversarial triplets and discriminative feature embedding in an unified framework. In particular, adversarial triplets are generated by introducing adversarial perturbations into the training process. This adversarial game is converted into a minimax problem so as to have an optimal solution from the theoretical view. Extensive experiments on several benchmark datasets demonstrate the effectiveness of the approach against the state-of-the-art literature.
\end{abstract}


\section{Introduction}
Person Re-Identification (ReID) learning has attracted much attention in the machine learning community \cite{alpher01,alpher32,alpher33}. It aims to judge whether two person images indicate the same target
or not and has widespread applications in video surveillance for public security. Inspired by the success of deep learning, many methods have been presented to deal with this task \cite{zheng2016person}. Among them, deep metric learning has become popular due to the seamless combination of the distance metric learning and deep neural network \cite{chengl15,alpher21,alpher16}. One representative work in the recent literature is TriNet \cite{hermans2017defense}, a convolutional neural network for learning embedding for person images. It exploits the triplet loss to learn an embedding space where the data points of the same identities are closer to each other than those with different identities. 

A key part of metric learning with triplet loss is the mining of hard triplets, since hard triplets in the training set will produce large gradients while the gradient of  easy triplets are close to zero. Although TriNet \cite{hermans2017defense} emphasizes the importance of hard example mining and propose a Batch-Hard mining strategy, which achieves competitive performance on person ReID benchmark, 
we observe that the loss of TriNet can still inevitably stagnate and confront with zero gradient. It is because that randomly chosen examples in a mini-batch still do not contain enough hard triplets, especially when the training dataset consists of a large number of distinct classes with few training instances for each, which is common for person ReID task. 
Since hard triplets usually account for a minority while most of the easy triplets makes little contribution to the optimization, a natural question is how to translate existing easy triplets into ``informative" hard cases so as to boost the generalization performance.

\begin{figure}
	\centering
	\includegraphics[width=0.5\linewidth]{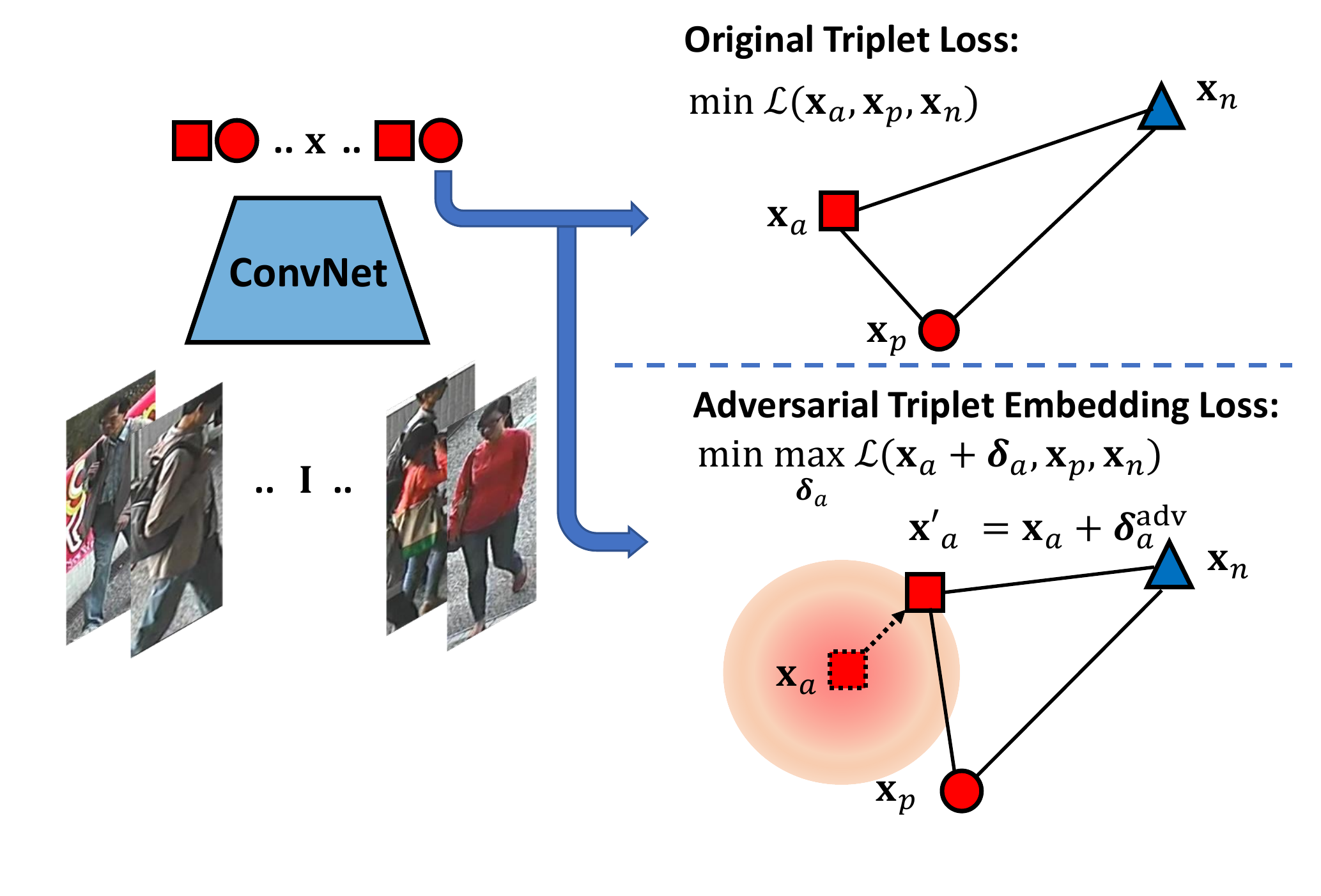}
	\caption{The comparison of traditional deep learning with (up) triplet loss and (bottom) adversarial triplet loss. Given a triplet $(\mathbf{x}_{a},\mathbf{x}_{p},\mathbf{x}_{n})$ consisting of an anchor $\vx_a$, its positive sample $\vx_p$ and negative sample $\vx_n$, whose labels satisfy $y_a = y_p \neq y_n$, triplet loss learns an embedding space where the data points of the same class are closer to each other than those with different classes. Our proposed ATE aims to introduce a bounded perturbation $\bm{\delta}_{a}$ into the anchor point $\mathbf{x}_{a}$ to generate a much harder triplet $(\mathbf{x}_{a}+\bm{\delta}_{a},\mathbf{x}_{p},\mathbf{x}_{n})$, in which the distance between $\vx_a$ and $\vx_p$ is enlarged while the distance between $\vx_a$ and $\vx_n$ is reduced. This mechanism is performed via an adversarial process and is further converted into a minimax problem to learn a worst case $\bm{\delta}_{a}^{adv}$ for the current loss. }
	\label{fig:illustrate}
	\vspace{-1.2em}
\end{figure}

To address the above problems, we propose a new metric learning objective called adversarial triplet embedding (ATE), which extend the triplet loss with an adversarial process. As shown in Fig{\ref{fig:illustrate}}, compared with the traditional traditional metric learning with the given triplet, our ATE learns to discriminate the generated adversarial triplets. Here, the adversarial triplets are automatically generated by introducing adversarial perturbations into the training process. Different from the traditional adversarial training \cite{goodfellowss14, miyato2016adversarial}, we apply the adversarial perturbation to the triplet loss, rather than directly to the input. The triplet loss model is pitted against an adversary: an adversarial mechanism that learns a worst case perturbation for the current loss. Further, we convert the adversarial game into a minimax problem which has the optimal solution from the theoretical perspective.

Note that the proposed adversarial manner is a good complements for the widely-used hard example mining. On one hand, it would provide a variety of adversarial triplets close to the margin through adding perturbations, which will improve the robustness of the model. On the other hand, the adversarial process would translate some easy triplets into hard ones which produce relatively large gradient for the optimization of the loss function. Consequently, through training with current hard example mining techniques, the proposed adversarial framework has a better generalization ability.

In experiment, we demonstrate the superiority of the proposed ATE to the triplet loss as well as other competing metric learning objectives on person ReID task. The ATE is evaluated on several real-world datasets. Extensive experiments on these benchmark datasets demonstrate the effectiveness of our approach against the state-of-the-art.

\section{Related Work}
The main topics of person ReID focus on discriminative feature learning and effective similarity measurement. For feature learning, many robust person image descriptors are developed for coping with misalignment and variations of color and texture. Commonly used hand-crafted descriptors include HSV color histogram \cite{alpher01}, local binary patterns (LBP) \cite{alpher03}, SIFT features \cite{alpher02}, and etc. Some efforts also exploit the properties of person images to boost the recognition rate, such as symmetry of local features \cite{alpher01} and the horizontal occurrence of local descriptors \cite{alpher32}. For similarity measurement, the key idea lies on metric learning which aims to learn an embedding from feature space to a new space so as to maximize the inter-class variations while minimizing the intra-class variations. Representative methods include local Fisher discriminant analysis \cite{alpher12}, Large Margin Nearest Neighbour \cite{alpher13}, KISS metric learning \cite{alpher14}, attribute consistent matching \cite{khamis2014joint}, and pair-wise constrained component analysis \cite{alpher11}.

Inspired by the success of deep learning networks, many efforts apply deep CNN models to the task of person re-identification and have achieved remarkable improvement over the approaches based on handcrafted features. In particular, end-to-end training \cite{alpher16, alpher21,wang2018resource} are exploited to learn discriminative representations and effective metrics simultaneously so that the feature learning network is directly optimized for the final task. Some methods considers person ReID as a ranking problem \cite{chengl15}, which is usually solved by deep networks with a triplet loss. Ding et al. \cite{alpher21} develop a triplet generation scheme for person re-identification by randomly select a small number of persons from the dataset. Cheng et al. \cite{alpher33} introduce a multi-channel parts-based CNN framework which improves the triplet loss function by imposes a margin constraint on the pair of matched images. Chen et al. \cite{alpher43} present a quadruplet loss for person re-identification, which extends the traditional triplet loss by considering pushing away negative pairs from positive pairs with different probe images.

In addition to the triple loss and its variants, the person ReID problem also can be tackled from the classification perspective \cite{alpher16, alpher22}. For example, Ahmed et al. \cite{alpher16} propose to learn representations and similarity metric simultaneously with an improved deep architecture, which takes pair-wise images as input and outputs a similarity value indicating whether the two input samples depict the same person. Zhang et al. \cite{yaqing2016} seek to learn cross-image representations for input image pairs and adopt the cross-entropy loss to represent the probability that the two images are of the same class (person) or not.

Further, triplet loss has shown its superiority on person re-identification task over other surrogate losses, such as classification and verification losses \cite{hermans2017defense}. A key part of learning with the triple loss is the mining of good hard triplets \cite{schroff2015facenet, shi2016embedding}. Batch Hard strategy has been proposed by Hermans et al. \cite{hermans2017defense} and has achieved state-of-the-art performance on several benchmark datasets. The core idea is to form batches by randomly sampling several person identities and then randomly sampling several images from each identity.

Our model differs from the above deep network with triplet loss. More specifically, we propose to train the triplet embedding with an adversarial process. A worst case perturbation for anchor point is learned to construct a much harder optimization problem, which encourages further reducing the intra-class variations and enlarging the inter-class variations. In particular, the learned perturbation help form an adaptive margin for triplet loss, which makes it possible to learn robust more representations for the person re-identification problem.

\section{The Proposed Method}
In this section, we present a new deep metric learning method called adversarial triplet embedding, where an adversarial perturbation is added to the anchor point to produce much harder triplets. We first overview the probabilistic interpretation of triplet embedding, and then investigate the statistical property of triplet embedding with perturbations. Further, we propose the adversarial triplet embedding through introducing adversarial perturbations into the training process.

\subsection{Probabilistic Triplet Embedding}
Suppose that we are provided with a set of training images  of $K$ classes, we first extract the feature embedding of each sample by CNN and obtain $\{(\vx, y), \cdots \}$, where $\vx \in \real^D$ denotes the feature embedding and $y \in \{1 ,\cdots ,K\}$ is the corresponding label.
At each training iteration, we sample a mini-batch of triplets, each of which $\cT=(\vx_a, \vx_p, \vx_n)$ consists of an anchor $\vx_a$, its positive sample $\vx_p$ and negative sample $\vx_n$, whose labels satisfy $y_a = y_p \neq y_n$. For the purpose of simplicity, we will explain the sampling strategy in Sec.~\ref{sec:detail}.
Triplet loss aims at learning a representation with correct ranking order, \ie, the distance between the anchor and the positive should less than the distance between the anchor and the negative:
\begin{equation}
	\norm{\vx_a - \vx_p }_2^2 <  \norm{\vx_a - \vx_n } _2^2.   \label{eq:constraint}
\end{equation}

Further, the corresponding probability of a given triplet \cite{Maaten2012StochasticTE} satisfying the above constraint can be written as:
\begin{align}
	p(\mathbf{x}_{a}, \mathbf{x}_{p}, \mathbf{x}_{n}|\phi)= & \frac{\exp(-\norm{\vx_a - \vx_p }_2^2 )}{\exp(-\norm{\vx_a - \vx_p }_2^2)+\exp(-\norm{\vx_a - \vx_n }_2^2)}\notag \\
	=                                                       & \frac{1}{1+\exp(\norm{\vx_a - \vx_p }_2^2-\norm{\vx_a - \vx_n }_2^2)}
\end{align}
In our case, $\mathbf{x}$ is deep representation with CNN. To learn the network parameters $\phi$ from a given set of triplets, we solve the following objective:
\begin{align}
	\arg\min_{\phi}\sum_{(\mathbf{x}_{a}, \mathbf{x}_{p}, \mathbf{x}_{n})}-\text{log}(p(\mathbf{x}_{a}, \mathbf{x}_{p}, \mathbf{x}_{n}|\phi))
\end{align}
the above objective can be interpreted as maximizing the likelihood in Eq.(\ref{eq:constraint}).

\subsection{Triplet Embedding with Perturbations}
Consider a triplet $(\mathbf{x}_{a}, \mathbf{x}_{p}, \mathbf{x}_{n})$, where $\mathbf{x}_{a}$ is corrupted with perturbations. Let $\tilde{\mathbf{x}}_{a}$ be the original uncorrupted anchor point. We assume that the triplet is generated as follows: first $(\tilde{\mathbf{x}}_{a}, \mathbf{x}_{p}, \mathbf{x}_{n})$ is generated according to a distribution $p(\tilde{\mathbf{x}}_{a}, \mathbf{x}_{p}, \mathbf{x}_{n} |\theta)$, where $\theta$ is an unknown parameter and would be estimated from the data set; given $(\tilde{\mathbf{x}}_{a}, \mathbf{x}_{p}, \mathbf{x}_{n})$, $\mathbf{x}_{a}$ is assumed to be generated from $\tilde{\mathbf{x}}_{a}$ according to a distribution $p(\mathbf{x}_{a}|\tilde{\theta}, \sigma_{a}, \tilde{\mathbf{x}}_{a})$, where $\tilde{\theta}$ is another unknown parameter and $\sigma_{a}$ is an estimated parameter for the perturbations of $\mathbf{x}_{a}$. The joint probability of $(\mathbf{x}_{a}, \tilde{\mathbf{x}}_{a}, \mathbf{x}_{p}, \mathbf{x}_{n})$ can be formulated as follows:
\begin{align}
	p(\mathbf{x}_{a}, \tilde{\mathbf{x}}_{a}, \mathbf{x}_{p}, \mathbf{x}_{n})=p(\tilde{\mathbf{x}}_{a}, \mathbf{x}_{p}, \mathbf{x}_{n} |\theta)p(\mathbf{x}_{a}|\tilde{\theta}, \sigma_{a}, \tilde{\mathbf{x}}_{a})
\end{align}
The joint probability distribution of $p(\mathbf{x}_{a}, \mathbf{x}_{p}, \mathbf{x}_{n})$ is computed by integrating out the unobserved quantity $\tilde{\mathbf{x}}_{a}$
\begin{align}
	p(\mathbf{x}_{a}, \mathbf{x}_{p}, \mathbf{x}_{n})=\int p(\tilde{\mathbf{x}}_{a}, \mathbf{x}_{p}, \mathbf{x}_{n} |\theta)p(\mathbf{x}_{a}|\tilde{\theta}, \sigma_{a}, \tilde{\mathbf{x}}_{a})d\tilde{\mathbf{x}}_{a}
\end{align}
This distribution can be considered as a probabilistic mixture model where each mixture component indicates a possible true anchor point $\mathbf{x}_{a}$. Further, the related parameters $(\theta, \tilde{\theta})$ can be estimated from the data through the maximum-likelihood estimation:
\begin{align}
  \max_{\theta, \tilde{\theta}} \sum_{(a,p,n)}\text{ln}p(\mathbf{x}_{a}, \mathbf{x}_{p}, \mathbf{x}_{n}|\theta, \tilde{\theta})
  =
  \max_{\theta, \tilde{\theta}}\sum_{(a,p,n)}\text{ln}\int p(\tilde{\mathbf{x}}_{a}, \mathbf{x}_{p}, \mathbf{x}_{n} |\theta)p(\mathbf{x}_{a}|\tilde{\theta}, \sigma_{a}, \tilde{\mathbf{x}}_{a})d\tilde{\mathbf{x}}_{a}
	\label{int}
\end{align}
However, this approach usually is difficult to solve as the integration over the unknown $\mathbf{x}_{a}$ has a very complicated formulation. Further, it is not easily extended to discriminative formulations such as maximum margin methods. Consequently, we consider an alternative approximation which is more tractable and usually employed in engineering applications. In particular, each $\tilde{\mathbf{x}}_{a}$ is regarded as a parameter of the probability distribution and thus the maximum-likelihood can be formulated as:
\begin{align}
	\max_{\theta,\tilde{\theta}}\sum_{(a,p,n)}\text{ln}\sup_{\tilde{\mathbf{x}}_{a}}\left[p(\tilde{\mathbf{x}}_{a}, \mathbf{x}_{p}, \mathbf{x}_{n} |\theta)p(\mathbf{x}_{a}|\tilde{\theta}, \sigma_{a}, \tilde{\mathbf{x}}_{a})\right]
	\label{discrete}
\end{align}

Since large values of $p(\tilde{\mathbf{x}}_{a}, \mathbf{x}_{p}, \mathbf{x}_{n} |\theta)p(\mathbf{x}_{a}|\tilde{\theta}, \sigma_{a}, \tilde{\mathbf{x}}_{a})$ decide the value of $\int p(\tilde{\mathbf{x}}_{a}, \mathbf{x}_{p}, \mathbf{x}_{n} |\theta)p(\mathbf{x}_{a}|\tilde{\theta}, \sigma_{a}, \tilde{\mathbf{x}}_{a})d\tilde{\mathbf{x}}_{a}$,  both objectives in Eq.(\ref{int}) and Eq.(\ref{discrete}) behave similarly and prefer to make the product $p(\tilde{\mathbf{x}}_{a}, \mathbf{x}_{p}, \mathbf{x}_{n} |\theta)p(\mathbf{x}_{a}|\tilde{\theta}, \sigma_{a}, \tilde{\mathbf{x}}_{a})$ large for some $\tilde{\mathbf{x}}_{a}$. On one hand, if an observation $\mathbf{x}_{a}$ is corrupted with large perturbations, we would select an appropriate $\tilde{\mathbf{x}}_{a}$ which predicts the probability well. On the other hand, if an observation $\mathbf{x}_{a}$ is corrupted with very small perturbations, then Eq.(\ref{int}) and Eq.(\ref{discrete}) would optimize the parameter of the model $\theta$ so that the $p(\tilde{\mathbf{x}}_{a}, \mathbf{x}_{p}, \mathbf{x}_{n} |\theta)p(\mathbf{x}_{a}|\tilde{\theta}, \sigma_{a}, \tilde{\mathbf{x}}_{a})$ is enough large. Such way of modeling would rely on data which are less corrupted and ignore those uncertain ones.

In conditionally probabilistic modeling, we assume that $p(\tilde{\mathbf{x}}_{a}, \mathbf{x}_{p}, \mathbf{x}_{n} |\theta)=p(\tilde{\mathbf{x}}_{a})p(\mathbf{x}_{p}, \mathbf{x}_{n}|\theta,\tilde{\mathbf{x}}_{a})$. As an example, we consider modeling with Gaussian perturbations,
\begin{align}
	p(\tilde{\mathbf{x}}_{a}, \mathbf{x}_{p}, \mathbf{x}_{n} |\theta)    & \sim \frac{1}{1+\exp(\norm{\tilde{\vx}_a - \vx_p }_2^2-\norm{\tilde{\vx}_a - \vx_n }_2^2)}\notag \\
	p(\mathbf{x}_{a}|\tilde{\theta}, \sigma_{a}, \tilde{\mathbf{x}}_{a}) & \sim \exp\left(-\frac{\norm{\mathbf{x}_{a}-\tilde{\mathbf{x}}_{a}}^2}{2\sigma_{a}^{2}}\right)
	\label{example}
\end{align}

The framework in Eq.(\ref{discrete}) becomes
\begin{align}
	\theta=\arg\min_{\theta}\sum_{(a,p,n)}\inf_{\tilde{\mathbf{x}}_{a}}\left[\text{ln}(1+L)+\frac{\norm{\mathbf{x}_{a}-\tilde{\mathbf{x}}_{a}}^2}{2\sigma_{a}^{2}}\right]
\end{align}
where $L=\exp(\norm{\tilde{\vx}_a - \vx_p }_2^2-\norm{\tilde{\vx}_a - \vx_n }_2^2)$.

\subsection{Adversarial Triplet Embedding}
Our formulation of adversarial triplet embedding (ATE) is motivated by adversarial examples training \cite{goodfellowss14, miyato2016adversarial}, which improves the robustness of classifier with adversarial perturbations on the input. Adversarial triplet embedding applies the adversarial perturbation to the triplet loss, rather than directly to the input. It is straightforward to put into use when $\mathbf{x}_{a}$ and $\mathbf{x}_{p}$ are learned with deep neural networks.

We assume that anchor points are subject to an additive perturbation, i.e., $\mathbf{x}_{a}=\mathbf{x}_{a}+\bm{\delta}_{a}$ where perturbation $\bm{\delta}_{a}$ follows certain distribution. In real-applications, bounded perturbations are usually discussed in adversarial training \cite{goodfellowss14, miyato2016adversarial} and the resulted methods exhibit robustness to the corrupted inputs. Thus, instead of adopting Gaussian noise as in Eq.(\ref{example}), we consider a simple bounded perturbation model $\norm{\bm{\delta}_{a}}\leq\epsilon_{a}$, which has a similar effect of the Gaussian noise model.

To maximize the effect of this bounded perturbation, we propose to train a triplet embedding function via an adversarial process, in which we simultaneously learn two components: a discriminative embedding space where the distance of samples with different classes is larger than that of samples with the same class, and an adversarial perturbation which increases the difficulty of the learning task as much as possible. This framework behaves like a minimax two-player game with the following loss function:
\begin{align}
	\min_{\theta}\sum_{(a,p,n)}\max_{\norm{\bm{\delta}_{a}}\leq \epsilon} & \left[\text{ln}(1+\exp(\norm{{\vx}_a +\bm{\delta}_{a}- \vx_p }_2^2\right.\notag \\
	                                                                      & \left.-\norm{{\vx}_a +\bm{\delta}_{a} - \vx_n }_2^2))\right]
	\label{ATEloss}
\end{align}
where the adversarial perturbation is introduced into the anchor point so that we reconstruct an more difficult problem, for which we would learn better similarity metric.

\subsection{Optimization and Extensions}
For each step of training, we learn the worst case perturbations $\bm{\delta}_{a}^{adv}$ against the current loss in Eq.(\ref{ATEloss}), and train the network to be robust to such perturbations through minimizing the following problem with respect to the network parameter $\theta$,
\begin{align}
	\cL_{\textrm{tri}} (\vx_a+\bm{\delta}_{a}^{adv},\vx_p, \vx_n;\theta) & =\left[\text{ln}(1+\exp(\norm{{\vx}_a +\bm{\delta}_{a}^{adv}- \vx_p }_2^2\right.\notag \\
	                                                                     & \left.-\norm{{\vx}_a +\bm{\delta}_{a}^{adv} - \vx_n }_2^2))\right]
	\label{adv_loss1}
\end{align}
where the adversarial perturbation $\bm{\delta}_{a}^{adv}$ would be computed as follows:
\begin{align}
	\bm{\delta}_{a}^{adv} & =\arg \max\limits_{\bm{\delta}_{a},\norm{\bm{\delta}_{a}}\leq\epsilon_a}\cL_{\textrm{tri}}(\vx_a+\bm{\delta}_{a},\vx_p,\vx_n;\hat{\theta})\notag \\
	                      & =\arg \max\limits_{\bm{\delta}_{a},\norm{\bm{\delta}_{a}}\leq\epsilon_a}\left[\text{ln}(1+\exp(\norm{{\vx}_a - \vx_p }_2^2\right. \notag         \\
	                      & \left.-\norm{{\vx}_a  - \vx_n }_2^2+2\bm{\delta}_{a}(\mathbf{x}_{n}-\mathbf{x}_{p})))\right]
	\label{adv_loss2}
\end{align}
where $\hat{\theta}$ uses a constant copy of the current parameters of the network. Such setup indicates that the backpropagation would not be used to update gradient through the adversarial loss construction process. With the Cauchy-Schwarz inequality, we have
\begin{align}
	\left|\bm{\delta}_{a}(\mathbf{x}_{n}-\mathbf{x}_{p})\right|\leq ||\bm{\delta}_{a}|| \cdot ||\mathbf{x}_{n}-\mathbf{x}_{p}||
\end{align}
where the equality holds if and only if $\bm{\delta}_{a}=k(\mathbf{x}_{n}-\mathbf{x}_{p})$ for certain scalar $k$. Based on the bounded constraint $\norm{\bm{\delta}_{a}}\leq\epsilon_a$, the worst case perturbation is obtained as follows:
\begin{align}
	\bm{\delta}_{a}^{adv}=\epsilon_{a}\frac{\mathbf{x}_{n}-\mathbf{x}_{p}}{||\mathbf{x}_{n}-\mathbf{x}_{p}||}
\end{align}
Consequently, the problem in Eq.(\ref{adv_loss1}) can be formulated as:
\begin{align}
	\cL_{\textrm{tri}} (\vx_a+\bm{\delta}_{a}^{adv},\vx_p, \vx_n;\theta) & =\left[\text{ln}(1+\exp(\norm{{\vx}_a - \vx_p }_2^2\right.\notag                       \\
	                                                                     & \left.-\norm{{\vx}_a - \vx_n }_2^2+s||\mathbf{x}_{n}-\mathbf{x}_{p}||_{2}^{2}))\right]
\label{adv_loss3}	                                                                  
\end{align}
where $s=2\epsilon_{a}/||\mathbf{x}_{n}-\mathbf{x}_{p}||$.

\noindent\textbf{Relationship to triplet loss} 
As in \cite{alpher33, alpher43}, the traditional triplet embedding seeks to minimize the following hinge loss:
\begin{equation}
	\cL_{\textrm{tri}} (\cT) = \left[\norm{\vx_a - \vx_p}_2^2
		-  \norm{\vx_a-\vx_n}_2^2 +m  \right]_+
	\label{eq:loss}
\end{equation}
where the operator $[\cdot ]_+ = \max (0, \cdot ) $ denotes the hinge function, which is used to avoid correcting "already correct " triplets. However, in person ReID it is necessary to pull together samples from the same identity as much as possible so as to reduce the intra-class variations. Based on this consideration, TriNet \cite{hermans2017defense} replaces the hinge function by a smooth version using the softplus function:  
\begin{align}
\cL_{\textrm{tri}} (\cT)=\left[\text{ln}(1+\exp(\norm{{\vx}_a - \vx_p }_2^2-\norm{{\vx}_a - \vx_n }_2^2))\right]
\end{align}
As shown in Eq.(\ref{adv_loss3}), this soft margin version is a part of our adversarial triplet loss, but without the third term in the $\exp(\cdot)$. The third term provides a help from the perspective of adaptive margin. It can further reduce the intra-class variations and enlarge the inter-class variations so that the generalization performance would be improved.

\subsection{Implementation Details} \label{sec:detail}

For batch construction during training we leverage the
idea of PK batches also introduced by \cite{hermans2017defense}.
This approach has shown very good performance in similarity-based ranking and avoids the need to generate a combinatorial number of triplets.
In each batch there are $K$ sample images for each of $P$ identities.
During one training epoch, each identity is selected in its batch in turn, and the remaining $P-1$ batch identities are sampled at random. $K$ samples for each identity are then also selected at random.

For sampling strategy within a batch, we modified the most hard sample strategy slightly to stabilize training process inspired by \cite{ristani2018features}.
Originally, the batch-hard sampling strategy consider only the
hardest positive and negative samples within a batch.
Compared to triplet loss without any sampling strategy, the batch-hard strategy converge faster and better, since it avoid the gradient of easy samples washing out the gradients of informative samples.
However, this method is sensitive to outliers and requires careful tunning of hyperparameter.
To overcome this weakness, we apply the softmax to the negative distance, obtain the importance of this sample in a probabilistic manner, and sample the positive and negative stochastically.




\section{Experiments}

In this section, we evaluate the ATE loss on the Person ReID task. Our method has been shown to achieve state-of-the-art performance on three public benchmark datasets.

\subsection{Dataset and Protocol}

We conduct experiments on three public benchmark datasets: CUHK03 \cite{li2014deepreid}, Market1501 \cite{zheng2015scalable},  VIPeR~\cite{Alpher29}, and PRID450s~\cite{Alpher38}.
The statistics of the datasets are shown in Tab.~\ref{tab:stat}. 

\begin{table}
  \centering 
	\caption{The statistics of three benchmark dataset}
	\label{tab:stat}
		\begin{tabular}[]{cccccccc}
			\toprule
			Dataset       & Market1501 & CUHK03    & VIPeR & PRID450s \\
			\midrule
			Identities    & 1501       & 1360      & 632   & 450 \\
			BBoxes        & 32,668     & 13,164    & 1264  & 900 \\
			Cameras       & 6          & 6         & 2     & 2 \\
			Label method  & DPM        & Hand/DPM  &   Hand    & Hand\\
			Train \# imgs & 12,936     & 7368/7365 & 632   &  450\\
			Train \# ids  & 751        & 767       & 316   & 225 \\
			Test \# imgs  & 19,732     & 1,400     & 632   &  450 \\
			Test \#ids    & 750        & 700       & 316   & 225 \\
			\bottomrule
    \end{tabular}
\end{table}

{\bf CUHK03} dataset contains 13,164 images of 1,360 identities.
It provides bounding boxes detected from deformable part models (DPMs) and manual labeling.
The traditional protocol split  the dataset into a training set containing 1,160 identities and a testing set containing 100 identities.
The  new training/testing protocol raised by \cite{zhong2017reranking} using a new training/testing protocol similar to that of Market-1501. The new protocol splits the dataset into training set consisting of 767 identities and testing set of 700 identities.
In testing, the new protocol randomly selects one image from each camera as a query for each identity and use the rest of images to construct the gallery set.
The new protocol gains two advantages: 1) For each identity, there are multiple ground truths in the gallery, which is more consistent with practical application scenario. 2) Evenly dividing the dataset into training set and testing set at once helps avoid time-consumingly repeating training and testing multiple times.

	{\bf Market1501}  dataset contains 32,668 images of 1,501 labeled
persons of six camera views. There are 751 identities in the training set and 750 identities in the testing set. In the original study on this proposed dataset, the author also uses mAP as the evaluation criteria to test the algorithms.


	{\bf VIPeR} dataset contains 632 person images captured by two cameras in an outdoor
environment, and each person has only one image in each
camera view.  
 
	{\bf PRID450s} dataset, similarly,  consists  of 450 identities,  both  captured
by two disjoint cameras.  
The widely adopted experimental protocol on two datasets is similar to the old protocol of cuhk03.
A random selection of half persons is used for training and the rest for testing. The
procedure is repeated for 10 times, then the average performances are reported. 
This procedure is time-consuming but acceptable on the small dataset, thus we strictly follow this widely adopted protocol. 

For evaluation metrics, we use several standard evaluation metrics: rank-n Cumulative Matching Characteristic accuracy,
which is an estimation of finding the corrected top-n match,
and mean average precision (mAP)
, which measures the overall quality of predicted ranklist. We use officially provided evaluation code for all the results. The only modifications are to use mean pooling on the embeddings of a tracklet rather than max pooling on Mars and to ignore the 6 queries with no right results in gallery.

\subsection{Experiments Settings}

We adpot the same settings as \cite{hermans2017defense} and \cite{wang2018resource}. We define an epoch as iterating over the whole dataset and train the network for 65 epochs. In this way, we do not need to tuning the number of iterations  for different datasets.
At training stage, Each input image is randomly cropped by a bounding box with random aspect ratio between $[1.5,3]$ and random area size between $[0.85,1]$, then randomly flipped horizontally, and resize to $256 \times 128$.
At test stage we resize the image to $256 \times 128$ and
average the embedding of the original image and of the horizontally flipped image to obtain the final embedding.
We take the ResNet-50 \cite{he2016deep}  pre-trained on ImageNet \cite{deng2009imagenet} as backbone network to extract features.
We train the model with Adam optimizer and batch size of 128, which contains 64 different people  and 4 different images each. The learning rate $\alpha$ is adjusted similarly as \cite{hermans2017defense}, starting from $\alpha_0 = 3 \times 10^{-4}$.
\begin{equation}
	\alpha (t)  =
	\begin{cases}
		\alpha_0                                      & \text{ if } t \le t_0,          \\
		\alpha_0 \times 0.001^{\frac{t-t_0}{t_1-t_0}} & \text { if } t_0 \le t \le t_1,
	\end{cases}
\end{equation}
where we set $t_0=35$ and $t_1=65$. $\beta_1$ of Adam is reduced from 0.9 to 0.5 after $t_0$ as well. All hyperparameters were taken from the optimized model of \cite{hermans2017defense}. Potentially we could improve the performance further through proper hyperparameter tuning.

%

\subsubsection{Baselines}

To demonstrate how ATE improves the generalization performance in comparison with the state-of-the-art person ReID methods, we compare it with the recent literature, including
PAN~\cite{zheng2018pedestrian}, SVDNet~\cite{sun2017svdnetfp},
LOMO + XQDA~\cite{alpher32}, LOMO + Null Space~\cite{zhang2016learning}, DTL~\cite{geng2016deep},
ResNet50 (I+V)~\cite{zheng2017discriminatively}, Gated siamese CNN~\cite{varior2016gated},
CNN + DCGAN~\cite{zheng2017unlabeled}, BraidNet-CS + SRL~\cite{wang2018person},
JLML~\cite{li2017person},
KISSME~\cite{alpher14},
LSSCDL~\cite{Alpher31},
ImprovedDL~\cite{alpher16},
and TCP~\cite{alpher33}. In addition, we also compare it with the following
triplet loss methods and its variants (with the same batch hard mining setup):
\begin{itemize}
	\item TriNet, a representative method for person ReID which exploits the triplet loss and proposes a batch hard mining scheme \cite{hermans2017defense}.
	\item Improved Triplet loss (ImpTrpLoss), which improves the triplet loss function by imposes a margin constraint on the pair of the same class \cite{alpher33}.
	\item QuadrupletNet, a model which extends the traditional triplet loss by considering pushing away negative pairs from positive pairs with different probe images \cite{alpher43}.
	\item TriNet+adversarial training (TriNetAdv), a model which introduces adversarial examples regularization to the triplet loss by making small perturbations to the input \cite{goodfellowss14}.
\end{itemize}

\subsection{Performance Comparison}
\textbf{Comparisons on CUHK03 dataset}.
We conduct the experiments on both labelled and detected CUHK03 datasets. From Table \ref{table:cuhk03}, we see that our proposed approach achieves the better results than the competing methods. On the labelled dataset, our method perform better than the next best method by an improvement of 3.03\% (53.03\% vs. 56.06\%) in the metric of mAP. On the detected dataset, the performance decreases a little due to the misalignment and incompleteness caused by the detector. However, the proposed method still achieves an improvement 3.41\% over the next best method (51.82\% vs. 55.23\%).

\textbf{Comparisons on Market1501 dataset}. 
From Table \ref{tab:mkt}, we see that our proposed method achieves the best performance of 86.49\% (rank-1) and 71.80\% (mAP) (vs. 85.12\% and 70.62\% respectively by the next best method). 

\textbf{Comparisons on VIPeR and PRID450s dataset}.
Following \cite{alpher16}, we pre-train the network using CUHK03 datasets, and fine-tune on the training set of VIPeR and PRID450s. As shown in the Table \ref{tab:viper}, the proposed ATE is better than the competing methods in all the cases except the rank-10 recognition rate for PRID450s dataset.

\begin{table}
	\centering
	\caption{Comparison of the state-of-the-art results on labelled and detected CUHK03 dataset (New protocol). The CMC scores (\%) at rank 1, 5, 10 and mAP are listed.} \label{table:cuhk03} 
	\vspace{1ex}
	\scalebox{.85}{
		\begin{tabular}{c|cc|cc}
			\hline
			\multirow{2}{*}{Methods}                               &
			\multicolumn{2}{c|}{ labelled CUHK03 }                 &
			\multicolumn{2}{c}{ detected CUHK03 } \cr
			\cline{2-3} \cline{4-5}
			                                                       & rank-1    & mAP       & rank-1    & mAP \cr
			\hline

			PAN \cite{zheng2018pedestrian}                         & 36.9      & 35.0      & 36.3      & 34.0      \\
			SVDNet   \cite{sun2017svdnetfp}                        & 40.9      & 37.8      & 41.5      & 37.2      \\
			IDE + XQDA       \cite{zhong2017reranking}             & 32.0      & 29.6      & 31.1      & 28.2      \\
			\hline \hline
			TriNet \cite{hermans2017defense}                       & 56.72     & 51.32     & 54.25     & 50.71     \\
			QuadrupletNet \cite{chen2017beyond}                    & 58.35     & 53.23     & 55.96     & 52.35     \\
			TriNetAdv \cite{goodfellow2014explainingah} & 54.66     & 49.71     & 51.19     & 48.02     \\
			ImpTrpLoss \cite{alpher33}                        & 57.98     & 53.03     & 55.23     & 51.82     \\
			ATE                                          & \bf 60.96 & \bf 56.06 & \bf 59.29 & \bf 55.23 \\
			\hline
		\end{tabular}
	}
\end{table}
%

\begin{table}
	\centering
	\caption{Comparison of the state-of-the-art results on Market1501 dataset.}
	\label{tab:mkt}
	\vspace{1ex}
	\scalebox{0.95}{
		\begin{tabular} {c|cc}
			\hline
			Methods                                                & rank-1    & mAP       \\ \hline
			LOMO + Null Space \cite{zhang2016learning}             & 55.43     & 29.87     \\
			DTL \cite{geng2016deep}                                & 83.7      & 65.6      \\
			ResNet50 (I+V) \cite{zheng2017discriminatively}        & 79.51     & 59.87     \\
			Gated siamese CNN \cite{varior2016gated}               & 65.88     & 39.55     \\
			CNN + DCGAN  \cite{zheng2017unlabeled}                 & 78.06     & 56.23     \\
			BraidNet-CS + SRL  \cite{wang2018person}               & 83.70     & 69.48     \\
			SVDNet \cite{sun2017svdnet}                            & 82.3      & 62.1      \\
			IDE+XQDA  \cite{zhong2017reranking}                    & 77.58     & 56.06     \\
			JLML  \cite{li2017person}                              & 85.1      & 65.5      \\ \hline \hline
			TriNet \cite{hermans2017defense}                       & 84.07     & 68.39     \\
			 QuadrupletNet \cite{chen2017beyond}                    & 85.12     & 70.62     \\
			TriNetAdv \cite{goodfellow2014explainingah} & 84.03     & 69.45     \\
			ImpTrpLoss \cite{alpher33}                        & 84.74     & 69.72     \\
			ATE                                          & \bf 86.49 & \bf 71.80     \\
			\hline
		\end{tabular}
	}
\end{table}


\begin{table}
	\centering
	\caption{Comparison of state-of-the-art results on VIPeR dataset. The cumulative matching scores (\%) at rank 1, 5, and 10 are listed.}
	\label{tab:viper}
	\vspace{1ex}
	\scalebox{0.75}{
		\begin{tabular}{c|ccc|ccc}
			\hline
			\multirow{2}*{Methods}                                 &
			\multicolumn{3}{c|}{VIPeR}                             &
			\multicolumn{3}{c}{PRID450s}                                                                                                        \\
			\cline{2-7}
			                                                       & r=1       & r=5        & r=10       & r=1        & r=5        & r=10       \\ \hline
			KISSME \cite{alpher14}                                 & 19.6      & 48.0       & 62.2       & 15.0       & -          & 39.0       \\
			LSSCDL \cite{Alpher31}                                 & 42.66     & -          & 84.27      & 60.49      & -          & 88.58      \\
			LOMO+XQDA  \cite{zhong2017reranking}                   & 40.00     & 68.13      & 80.51      & 61.42      & -          & 90.84      \\
			ImprovedDL \cite{alpher16}                             & 34.81     & 63.61      & 75.63      & 34.81      & 63.72      & 76.24      \\
			TCP \cite{alpher33}                                    & 47.80     & 74.70      & 84.80      & -          & -          & -          \\
			SSM \cite{bai2017}                                     & 53.73     & -          & 91.49      & 72.98      & -          & \bf 96.76  \\
			FFN \cite{wu2016}                                      & 51.06     & 81.01      & 91.39      & 66.62      & 86.84      & 92.84      \\
			Mirror-KMFA \cite{chen2015mirror}                      & 42.97     & 75.82      & 87.28      & 55.42      & 79.29      & 87.82      \\
                         \hline
			TriNet \cite{hermans2017defense}                       & 55.26     & 82.24      & 91.62      & 73.76      & 91.67      & 95.13      \\
			QuadrupletNet \cite{chen2017beyond}                    & 56.24     & 82.81      & 91.69      & 73.82      & 91.95      & \bf 95.23  \\
			TriNetAdv \cite{goodfellow2014explainingah} & 54.53     & 81.62      & 91.22      & 72.14      & 90.81      & 94.12      \\
			ImpTrpLoss \cite{alpher33}                        & 55.39     & 82.25      & 91.59      & 73.74      & 91.53      & 95.06      \\
			ATE                                          & \bf 56.45 & \bf  83.11 & \bf  91.78 & \bf  73.91 & \bf  92.03 & \bf  95.24 \\
                         \hline
		\end{tabular}
	}
\end{table}
\begin{figure*}
	\centering 
	\includegraphics[width=.99\linewidth]{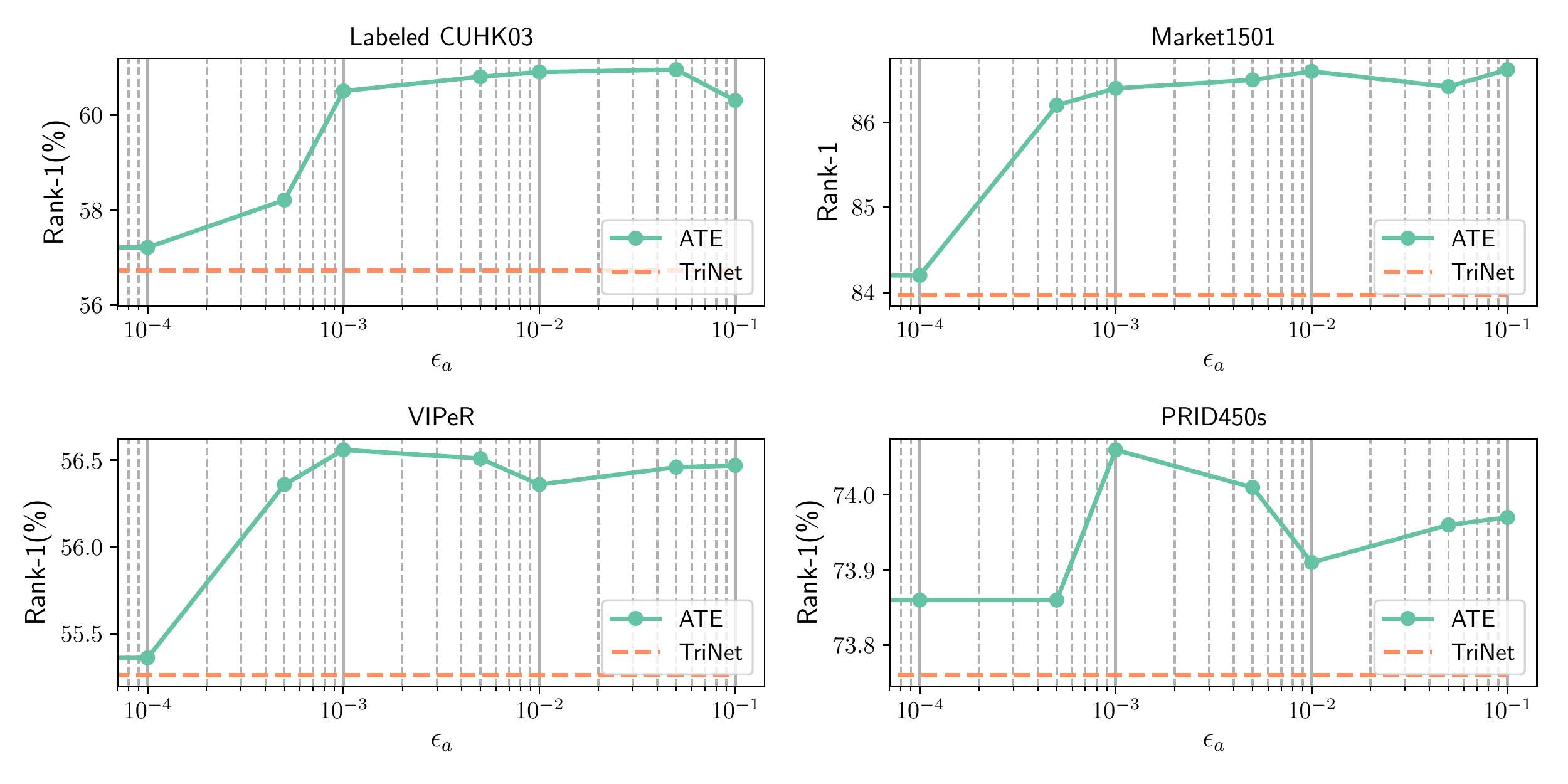}
	\caption{The rank-1 score for the testing set as a function of perturbation bound parameter $\epsilon_{a}$ for ATE on the labelled CUHK03, Market 1501, VIPeR, and PRID450s datasets, respectively.} \label{fig:hyperparam}
\end{figure*}
{\bf Triplet loss}: as shown in Table \ref{table:cuhk03}, \ref{tab:mkt}, and \ref{tab:viper}, TriNet achieves competitive performance in these datasets by emphasizing the importance of hard example mining for deep learning with triple loss. Compared with other traditional ReID approaches, triplet loss shows a significant benefit by directly performing end-to-end learning between the input images and the desired ranking relationship.

{\bf QuadrupletNet vs. TriNet}: as shown in Table \ref{table:cuhk03}, \ref{tab:mkt}, and \ref{tab:viper}, Quadruplet performs better than TriNet with the same setup of batch hard mining. This is because it extends the traditional triplet loss by considering pushing away negative pairs from positive pairs with different probe images, which provides a help from the aspect of orders with different probe images and further enlarge the inter-class variations.

{\bf ImpTrpLoss vs. TriNet}: as shown in Table \ref{table:cuhk03}, \ref{tab:mkt}, and \ref{tab:viper}, ImpTrpLoss performs a little better than TriNet in most cases. This is because it improves the triplet loss function by imposes a margin constraint on the pair of matched images, which reduces the intra-class variations and improve the performance on the test data.

{\bf TriNetAdv vs. TriNet}: as shown in Table \ref{table:cuhk03}, \ref{tab:mkt}, and \ref{tab:viper}, TriNetAdv does not show superiority over TriNet by introducing adversarial training regularization. Intuitively, the adversarial perturbation on pixel space is not semantically meaningful and not consistent to the distribution of test set, thus adversarial training do not contribute a lot to the improvement of performance and sometime even worse the accuracy on test set.

{\bf ATE vs. TriNet}: from Table \ref{table:cuhk03}, \ref{tab:mkt}, and \ref{tab:viper}, we see that ATE performs significantly better than TriNet. 
On one hand, this illustrates the importance of simultaneously considering learning adversarial triplets and discriminative feature embedding in an adversarial manner. 
The adversarial triplets generated by this adversarial manner are good complements for the widely-used hard example mining. 
On the other hand, the adaptive margin provided by the proposed model further enlarges the inter-class variations while reducing the intra-class variations, which improves the generalization performance. 
In particular, our proposed method outperforms TriNet by an improvement on rank-1 scores of 4.24\% (60.96\% v.s. 56.72\%) on labelled CUHK03 dataset and 5.04\% (59.29\% v.s. 54.25\%) on detected CUHK03 dataset. 
Similarly, on the Market1501 dataset our model also achieves an improvement over TriNet on rank-1score of 2.42\% (86.49\% v.s. 84.07). 

\begin{figure*}
	\centering
	\includegraphics[width=0.98\linewidth]{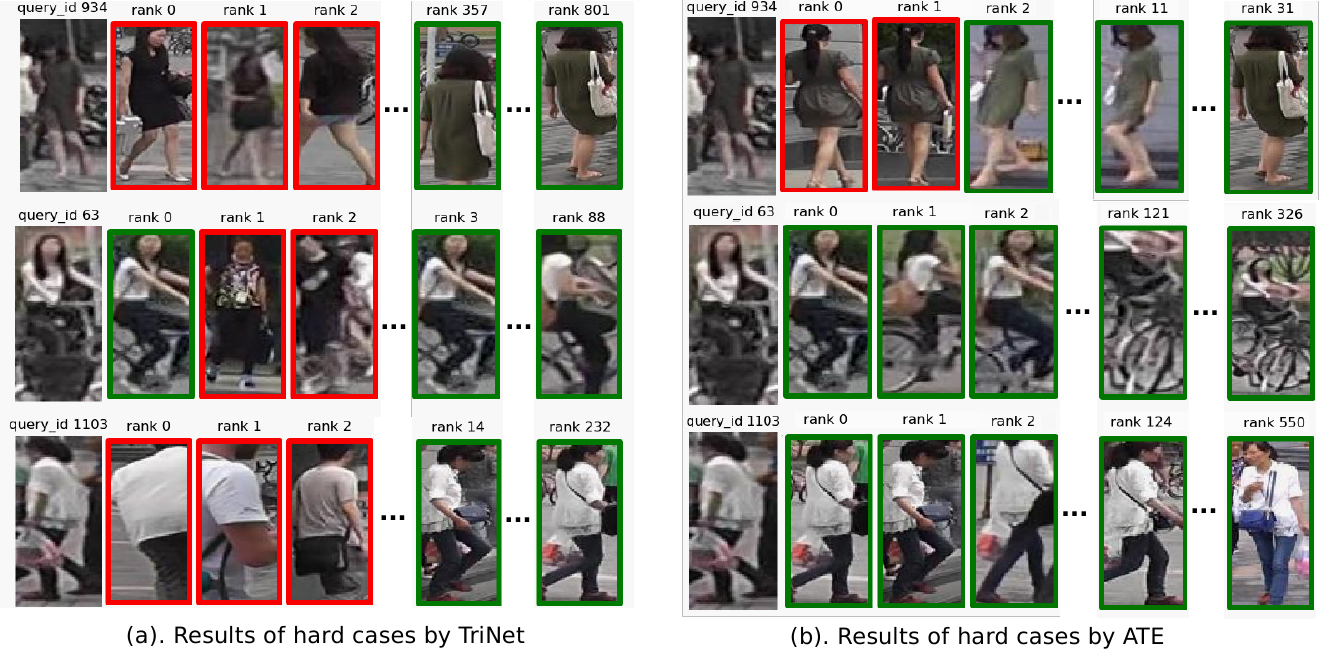}
	\caption{Visualization of the recognition results for hard cases of the Market1501 dataset by TriNet and our proposed model, respectively. }
	\label{fig:visual}
\end{figure*}
\subsection{Sensitivity Study}
We conduct a sensitivity study on how the bound parameter $\epsilon_{a}$ affect the recognition performance of the proposed model in terms of rank-1. Figure \ref{fig:hyperparam} shows how the parameter $\epsilon_{a}$ affects the recognition performance of the proposed model in terms of rank-1. This bound hyperparameter $\delta_a$ controls the strength of the adversarial perturbation. Too small a $\epsilon_a$ is selected, the ATE loss would make no difference with the Triplet loss. When too large a $\epsilon_a$ is selected, it would be unstable to optimize or stagnate  at local minimum and obtain a degraded performance. However, in practice, it is observed that ATE is not very sensitive to the selection of $\epsilon_a$ and produces fairly good improvement in a wide range of hyperparameter. To illustrate this point, we evaluate its performance at $\epsilon_a=[10^{-4}, 5\times 10^{-3}, 10^{-3}, 5 \times 10^{-2}, 10^{-2}, 5\times 10^{-1}, 10^{-1}]$. In Fig.~\ref{fig:hyperparam}, the red dashed line is TriNet baseline, and the blue line represents rank-1 of ATE model. We conclude $\epsilon_a=10^{-2}$ is the best for large dataset like CUHK03 and Market1501, $\epsilon_a=10^{-3}$ is the best for small dataset like VIPeR and PRID450s, and there is no great difference on rank-1 metric between selecting $10^{-3}$ and $10^{-2}$. We guess $\epsilon_a$ may slightly depended on the distribution and characteristic of datasets, \eg, a large dataset allows us to adopt a large adversarial perturbation to explore the optimal margin for learning discriminative features.

\subsection{Visualization of Hard Cases}
From the TriNet model to the ATE model on Market1501 Dataset, the mAP increase from 68.39\% to 71.80\%, by a margin of 2.77\%. There are 4.3\% queries which improves more than 75\% AP. These queries are poorly predicted by the model supervised by the original triplet loss, and greatly corrected by ATE loss, which shows the superiority of our model. Further, we rank the queries by its improvement of AP, and choose several successfully corrected cases for visualization. As shown in \ref{fig:visual}, the proposed ATE model is robust to the large cross-view appearance variations caused by mutual occlusions, background clutters, distractors and misalignment caused by detector, and doppelganger identities. It is contributed to the explicit model the adversarial perturbation in the feature space in training process. For example, in row one, the TriNet model is misled by a similar identity of query 934. The ATE model is still confused by a more similar but reasonable doppelganger identity, and rank the gallery image correct identity higher in the rank list. In row two, although the query image is cluttered by background and of low quality,  
the ATE model successfully retrieves both high-quality and low quality image. In row three, the TriNet model is confused by distractor caused by poor detector, which is the error of previous detection stage. The ATE model, however, is resistant to this error and successfully retrieve the correct images. 

\section{Conclusion}
In this paper, we develop a new deep metric learning method called Adversarial Triplet Embedding (ATE) for person ReID, in which we simultaneously generate adversarial triplets and discriminative feature embedding in an unified framework. In particular, adversarial triplets are generated by introducing adversarial perturbations into the training process. Further, this adversarial process is a good complements for the widely-used hard example mining. In addition, we convert this adversarial game into a minimax problem so as to have an optimal solution from the theoretical aspect. Extensive experiments on several benchmark datasets demonstrate the effectiveness of the approach against the state-of-the-art literature.

\bibliographystyle{unsrt}  
\bibliography{egbib}  

\begin{thebibliography}{10}

\bibitem{alpher01}
Michela Farenzena, Loris Bazzani, Alessandro Perina, Vittorio Murino, and Marco
  Cristani.
\newblock Person re-identification by symmetry-driven accumulation of local
  features.
\newblock In {\em CVPR}, pages 2360--2367, 2010.

\bibitem{alpher32}
Shengcai Liao, Yang Hu, Xiangyu Zhu, and Stan~Z Li.
\newblock Person re-identification by local maximal occurrence representation
  and metric learning.
\newblock In {\em CVPR}, pages 2197--2206, 2015.

\bibitem{alpher33}
De~Cheng, Yihong Gong, Sanping Zhou, Jinjun Wang, and Nanning Zheng.
\newblock Person re-identification by multi-channel parts-based cnn with
  improved triplet loss function.
\newblock In {\em CVPR}, pages 1335--1344, 2016.

\bibitem{zheng2016person}
Liang Zheng, Yi~Yang, and Alexander~G Hauptmann.
\newblock Person re-identification: Past, present and future.
\newblock {\em arXiv preprint arXiv:1610.02984}, 2016.

\bibitem{chengl15}
Shi{-}Zhe Chen, Chun{-}Chao Guo, and Jian{-}Huang Lai.
\newblock Deep ranking for person re-identification via joint representation
  learning.
\newblock {\em CoRR}, abs/1505.06821, 2015.

\bibitem{alpher21}
Shengyong Ding, Liang Lin, Guangrun Wang, and Hongyang Chao.
\newblock Deep feature learning with relative distance comparison for person
  re-identification.
\newblock {\em Pattern Recognition}, 48(10):2993--3003, 2015.

\bibitem{alpher16}
Ejaz Ahmed, Michael Jones, and Tim~K Marks.
\newblock An improved deep learning architecture for person re-identification.
\newblock In {\em CVPR}, pages 3908--3916, 2015.

\bibitem{hermans2017defense}
Alexander Hermans, Lucas Beyer, and Bastian Leibe.
\newblock In defense of the triplet loss for person re-identification.
\newblock {\em arXiv preprint arXiv:1703.07737}, 2017.

\bibitem{goodfellowss14}
Ian~J. Goodfellow, Jonathon Shlens, and Christian Szegedy.
\newblock Explaining and harnessing adversarial examples.
\newblock {\em CoRR}, abs/1412.6572, 2014.

\bibitem{miyato2016adversarial}
Takeru Miyato, Andrew~M Dai, and Ian Goodfellow.
\newblock Adversarial training methods for semi-supervised text classification.
\newblock {\em arXiv preprint arXiv:1605.07725}, 2016.

\bibitem{alpher03}
Wei Li and Xiaogang Wang.
\newblock Locally aligned feature transforms across views.
\newblock In {\em CVPR}, pages 3594--3601, 2013.

\bibitem{alpher02}
Rui Zhao, Wanli Ouyang, and Xiaogang Wang.
\newblock Unsupervised salience learning for person reidentification.
\newblock In {\em CVPR}, pages 3586--3593, 2013.

\bibitem{alpher12}
Sateesh Pedagadi, James Orwell, Sergio Velastin, and Boghos Boghossian.
\newblock Local fisher discriminant analysis for pedestrian re-identification.
\newblock In {\em CVPR}, pages 3318--3325, 2013.

\bibitem{alpher13}
Kilian~Q Weinberger and Lawrence~K Saul.
\newblock Distance metric learning for large margin nearest neighbor
  classification.
\newblock {\em Journal of Machine Learning Research}, 10(Feb):207--244, 2009.

\bibitem{alpher14}
Martin Koestinger, Martin Hirzer, Paul Wohlhart, Peter~M Roth, and Horst
  Bischof.
\newblock Large scale metric learning from equivalence constraints.
\newblock In {\em CVPR}, pages 2288--2295. IEEE, 2012.

\bibitem{khamis2014joint}
Sameh Khamis, Cheng-Hao Kuo, Vivek~K Singh, Vinay~D Shet, and Larry~S Davis.
\newblock Joint learning for attribute-consistent person re-identification.
\newblock In {\em European Conference on Computer Vision}, pages 134--146,
  2014.

\bibitem{alpher11}
Alexis Mignon and Fr{\'e}d{\'e}ric Jurie.
\newblock Pcca: A new approach for distance learning from sparse pairwise
  constraints.
\newblock In {\em CVPR}, pages 2666--2672, 2012.

\bibitem{wang2018resource}
Yan Wang, Lequn Wang, Yurong You, Xu~Zou, Vincent Chen, Serena Li, Gao Huang,
  Bharath Hariharan, and Kilian~Q Weinberger.
\newblock Resource aware person re-identification across multiple resolutions.
\newblock In {\em Proceedings of the IEEE Conference on Computer Vision and
  Pattern Recognition}, pages 8042--8051, 2018.

\bibitem{alpher43}
Weihua Chen, Xiaotang Chen, Jianguo Zhang, and Kaiqi Huang.
\newblock A multi-task deep network for person re-identification.
\newblock In {\em AAAI}, pages 3988--3994, 2017.

\bibitem{alpher22}
Faqiang Wang, Wangmeng Zuo, Liang Lin, David Zhang, and Lei Zhang.
\newblock Joint learning of single-image and cross-image representations for
  person re-identification.
\newblock In {\em CVPR}, pages 1288--1296, 2016.

\bibitem{yaqing2016}
Zhang Yaqing, Xi~Li, Liming Zhao, and Zhang Zhongfei.
\newblock Semantics-aware deep correspondence structure learning for robust
  person re-identification.
\newblock {\em IJCAI Int. Jt. Conf. Artif. Intell.}, 2016-Janua:3545--3551,
  2016.

\bibitem{schroff2015facenet}
Florian Schroff, Dmitry Kalenichenko, and James Philbin.
\newblock Facenet: A unified embedding for face recognition and clustering.
\newblock In {\em Proceedings of the IEEE conference on computer vision and
  pattern recognition}, pages 815--823, 2015.

\bibitem{shi2016embedding}
Hailin Shi, Yang Yang, Xiangyu Zhu, Shengcai Liao, Zhen Lei, Weishi Zheng, and
  Stan~Z Li.
\newblock Embedding deep metric for person re-identification: A study against
  large variations.
\newblock In {\em European Conference on Computer Vision}, pages 732--748,
  2016.

\bibitem{Maaten2012StochasticTE}
Laurens van~der Maaten and Kilian~Q. Weinberger.
\newblock Stochastic triplet embedding.
\newblock {\em 2012 IEEE International Workshop on Machine Learning for Signal
  Processing}, pages 1--6, 2012.

\bibitem{ristani2018features}
Ergys Ristani and Carlo Tomasi.
\newblock Features for multi-target multi-camera tracking and
  re-identification.
\newblock {\em arXiv preprint arXiv:1803.10859}, 2018.

\bibitem{li2014deepreid}
Wei Li, Rui Zhao, Tong Xiao, and Xiaogang Wang.
\newblock Deepreid: Deep filter pairing neural network for person
  re-identification.
\newblock In {\em Proceedings of the IEEE Conference on Computer Vision and
  Pattern Recognition}, pages 152--159, 2014.

\bibitem{zheng2015scalable}
Liang Zheng, Liyue Shen, Lu~Tian, Shengjin Wang, Jingdong Wang, and Qi~Tian.
\newblock Scalable person re-identification: A benchmark.
\newblock In {\em Proceedings of the IEEE International Conference on Computer
  Vision}, pages 1116--1124, 2015.

\bibitem{Alpher29}
Douglas Gray and Hai Tao.
\newblock Viewpoint invariant pedestrian recognition with an ensemble of
  localized features.
\newblock {\em Computer Vision--ECCV}, pages 262--275, 2008.

\bibitem{Alpher38}
Peter~M Roth, Martin Hirzer, Martin Koestinger, Csaba Beleznai, and Horst
  Bischof.
\newblock Mahalanobis distance learning for person re-identification.
\newblock In {\em Person Re-Identification}, pages 247--267. Springer, 2014.

\bibitem{zhong2017reranking}
Zhun Zhong, Liang Zheng, Donglin Cao, and Shaozi Li.
\newblock Re-ranking person re-identification with k-reciprocal encoding.
\newblock In {\em Computer Vision and Pattern Recognition (CVPR), 2017 IEEE
  Conference on}, pages 3652--3661. IEEE, 2017.

\bibitem{he2016deep}
Kaiming He, Xiangyu Zhang, Shaoqing Ren, and Jian Sun.
\newblock Deep residual learning for image recognition.
\newblock In {\em Proceedings of the IEEE conference on computer vision and
  pattern recognition}, pages 770--778, 2016.

\bibitem{deng2009imagenet}
Jia Deng, Wei Dong, Richard Socher, Li-Jia Li, Kai Li, and Li~Fei-Fei.
\newblock Imagenet: A large-scale hierarchical image database.
\newblock In {\em Computer Vision and Pattern Recognition, 2009. CVPR 2009.
  IEEE Conference on}, pages 248--255. Ieee, 2009.

\bibitem{zheng2018pedestrian}
Zhedong Zheng, Liang Zheng, and Yi~Yang.
\newblock Pedestrian alignment network for large-scale person
  re-identification.
\newblock {\em IEEE Transactions on Circuits and Systems for Video Technology},
  2018.

\bibitem{sun2017svdnetfp}
Yifan Sun, Liang Zheng, Weijian Deng, and Shengjin Wang.
\newblock Svdnet for pedestrian retrieval.
\newblock {\em 2017 IEEE International Conference on Computer Vision (ICCV)},
  pages 3820--3828, 2017.

\bibitem{zhang2016learning}
Li~Zhang, Tao Xiang, and Shaogang Gong.
\newblock Learning a discriminative null space for person re-identification.
\newblock In {\em Proceedings of the IEEE conference on computer vision and
  pattern recognition}, pages 1239--1248, 2016.

\bibitem{geng2016deep}
Mengyue Geng, Yaowei Wang, Tao Xiang, and Yonghong Tian.
\newblock Deep transfer learning for person re-identification.
\newblock {\em arXiv preprint arXiv:1611.05244}, 2016.

\bibitem{zheng2017discriminatively}
Zhedong Zheng, Liang Zheng, and Yi~Yang.
\newblock A discriminatively learned cnn embedding for person reidentification.
\newblock {\em ACM Transactions on Multimedia Computing, Communications, and
  Applications (TOMM)}, 14(1):13, 2017.

\bibitem{varior2016gated}
Rahul~Rama Varior, Mrinal Haloi, and Gang Wang.
\newblock Gated siamese convolutional neural network architecture for human
  re-identification.
\newblock In {\em European Conference on Computer Vision}, pages 791--808.
  Springer, 2016.

\bibitem{zheng2017unlabeled}
Zhedong Zheng, Liang Zheng, and Yi~Yang.
\newblock Unlabeled samples generated by gan improve the person
  re-identification baseline in vitro.
\newblock {\em arXiv preprint arXiv:1701.07717}, 3, 2017.

\bibitem{wang2018person}
Yicheng Wang, Zhenzhong Chen, Feng Wu, and Gang Wang.
\newblock Person re-identification with cascaded pairwise convolutions.
\newblock In {\em Proceedings of the IEEE Conference on Computer Vision and
  Pattern Recognition}, pages 1470--1478, 2018.

\bibitem{li2017person}
Wei Li, Xiatian Zhu, and Shaogang Gong.
\newblock Person re-identification by deep joint learning of multi-loss
  classification.
\newblock {\em arXiv preprint arXiv:1705.04724}, 2017.

\bibitem{Alpher31}
Ying Zhang, Baohua Li, Huchuan Lu, Atshushi Irie, and Xiang Ruan.
\newblock Sample-specific svm learning for person re-identification.
\newblock In {\em CVPR}, pages 1278--1287, 2016.

\bibitem{chen2017beyond}
Weihua Chen, Xiaotang Chen, Jianguo Zhang, and Kaiqi Huang.
\newblock Beyond triplet loss: a deep quadruplet network for person
  re-identification.
\newblock In {\em The IEEE Conference on Computer Vision and Pattern
  Recognition (CVPR)}, volume~2, 2017.

\bibitem{goodfellow2014explainingah}
Ian~J. Goodfellow, Jonathon Shlens, and Christian Szegedy.
\newblock Explaining and harnessing adversarial examples.
\newblock {\em CoRR}, abs/1412.6572, 2014.

\bibitem{sun2017svdnet}
Yifan Sun, Liang Zheng, Weijian Deng, and Shengjin Wang.
\newblock Svdnet for pedestrian retrieval.
\newblock {\em arXiv preprint}, 1(6), 2017.

\bibitem{bai2017}
Song Bai, Xiang Bai, and Qi~Tian.
\newblock Scalable person re-identification on supervised smoothed manifold.
\newblock {\em 2017 IEEE Conference on Computer Vision and Pattern Recognition
  (CVPR)}, Jul 2017.

\bibitem{wu2016}
Shangxuan Wu, Ying-Cong Chen, Xiang Li, An-Cong Wu, Jin-Jie You, and Wei-Shi
  Zheng.
\newblock An enhanced deep feature representation for person re-identification.
\newblock {\em 2016 IEEE Winter Conference on Applications of Computer Vision
  (WACV)}, Mar 2016.

\bibitem{chen2015mirror}
Ying-Cong Chen, Wei-Shi Zheng, and Jianhuang Lai.
\newblock Mirror representation for modeling view-specific transform in person
  re-identification.
\newblock In {\em IJCAI}, pages 3402--3408, 2015.

\end{thebibliography}






\end{document}